\documentclass[sigconf,natbib=true,anonymous=false]{acmart}

\renewcommand\footnotetextcopyrightpermission[1]{} 
\AtBeginDocument{%
  \providecommand\BibTeX{{%
    \normalfont B\kern-0.5em{\scshape i\kern-0.25em b}\kern-0.8em\TeX}}}

\usepackage{enumitem}

\begin{document}

\title{MACSA: A Multimodal Aspect-Category Sentiment Analysis Dataset with Multimodal Fine-grained Aligned Annotations}


\author{Hao Yang}
\email{hyang@ir.hit.edu.cn}
\affiliation{%
  \institution{Harbin Institute of Technology}
  \city{Harbin}
  \country{China}
}
\author{Yanyan Zhao}
\email{yyzhao@ir.hit.edu.cn}
\affiliation{%
  \institution{Harbin Institute of Technology}
  \city{Harbin}
  \country{China}
}
\author{Jianwei Liu}
\email{jwliu@ir.hit.edu.cn}
\affiliation{%
  \institution{Harbin Institute of Technology}
  \city{Harbin}
  \country{China}
}
\author{Yang Wu}
\email{ywu@ir.hit.edu.cn}
\affiliation{%
  \institution{Harbin Institute of Technology}
  \city{Harbin}
  \country{China}
}
\author{Bing Qin}
\email{qinb@ir.hit.edu.cn}
\affiliation{%
  \institution{Harbin Institute of Technology}
  \city{Harbin}
  \country{China}
}


\begin{abstract}
Multimodal fine-grained sentiment analysis has recently attracted increasing attention due to its broad applications. However, the existing multimodal fine-grained sentiment datasets most focus on annotating the fine-grained elements in text but ignore those in images, which leads to the fine-grained elements in visual content not receiving the full attention they deserve. In this paper, we propose a new dataset, the Multimodal Aspect-Category Sentiment Analysis (MACSA) dataset, which contains more than 21K 
text-image pairs. The dataset provides fine-grained annotations for both textual and visual content and firstly uses the aspect category as the pivot to align the fine-grained elements between the two modalities. Based on our dataset, we propose the Multimodal ACSA task and a multimodal graph-based aligned model (MGAM), which adopts a fine-grained cross-modal fusion method. Experimental results show that our method can facilitate the baseline comparison for future research on this corpus. We will make the dataset and code publicly available.
\end{abstract}

\maketitle

\pagestyle{empty}

\section{Introduction}

Inspired by textual fine-grained sentiment analysis, multimodal fine-grained sentiment analysis has attracted many researchers' attention. The existing works mainly focus on the multimodal Aspect-based Sentiment Analysis (ABSA) task, which aims to detect the sentiment towards a given aspect.
And the related works include Multi-ZOL dataset\cite{xu2019multi} with Multi-Interactive Memory Network (MIMN) and Twitter dataset\cite{yu2019adapting} with Target-oriented multimodal BERT (TomBERT). We observe that in these multimodal ABSA datasets, the researchers only annotate the fine-grained elements in the text, and there are no annotations for the equipped image. In other words, for a text-image pair, the ``fine-grained'' is just for the text but not for the image.

Actually, the fine-grained elements from text and image are both important and explainable clues. Ignoring either part will lead to deviation from the actual application. Moreover, as the saying goes, ``a picture is worth a thousand words'', the image also provides useful and important information in multimodal data as the textual content. Ignoring the fine-grained analysis of images will lead to deviation in the overall analysis of multimodal data. Thus the image is worth being fine-grained annotated. 

\begin{figure}[ht]
    \centering
    \includegraphics[width=8cm]{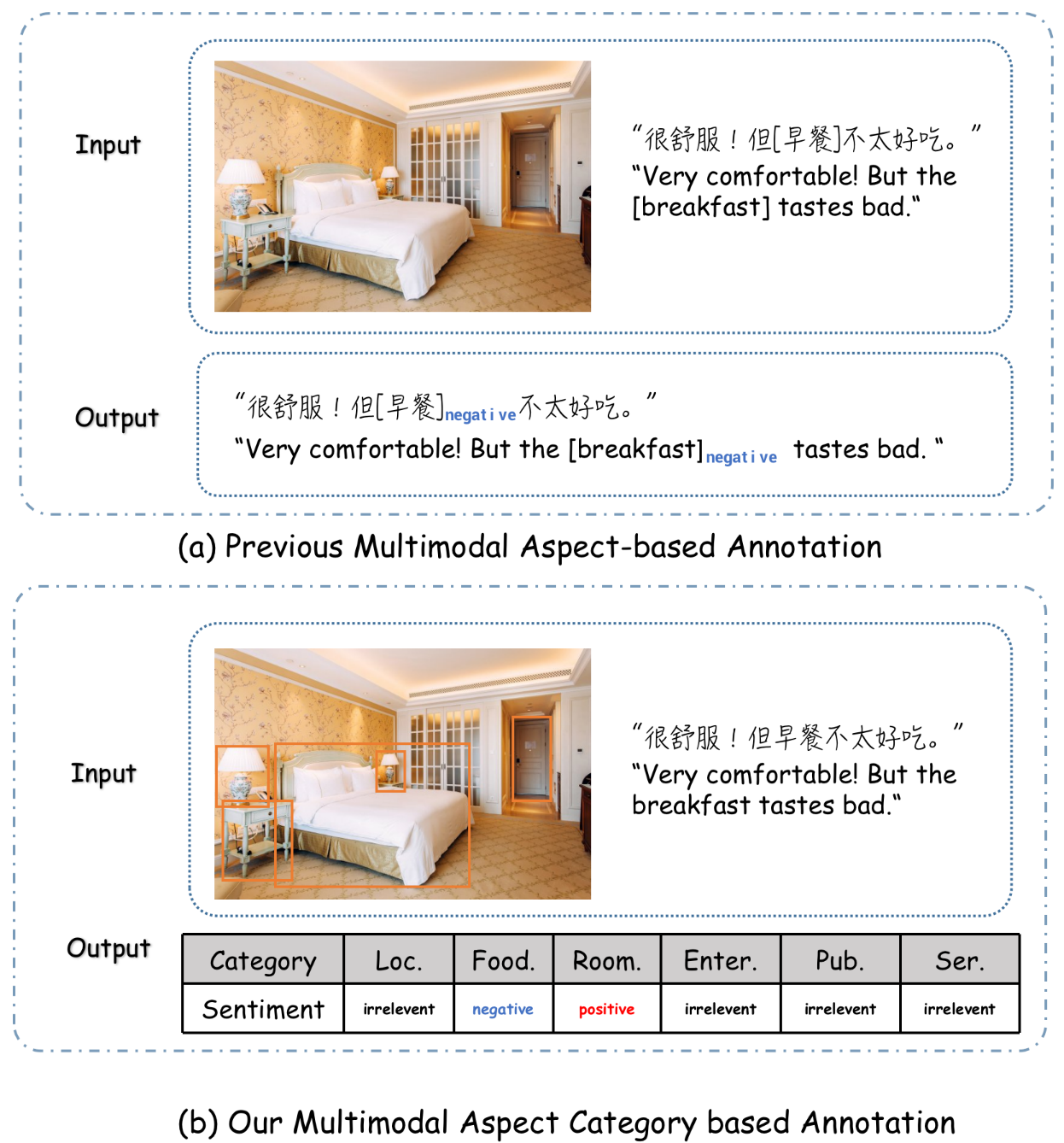}
    \caption{The comparison between the previous multimodal ABSA annotation/task and our ACSA annotation/task.}
    \label{fig-Example}
\end{figure}

This paper proposes to construct a multimodal dataset with textual and visual fine-grained annotations. This dataset uses the \textbf{``aspect category''} as a pivot to bridge the elements between the image and the text, and we name this dataset MACSA. Based on this dataset, we propose a new \textit{multimodal aspect-category sentiment analysis} task, which aims to identify the sentiment towards each pre-defined aspect category. As shown in Figure ~\ref{fig-Example}(b), the related categories for the text-image pair include ``\emph{Room}'' and ``\emph{Food}''. And the task is to first detect the related categories and then recognize the polarities for the related categories.
We compare the previous multimodal ABSA annotation/task and our new ACSA annotation/task in Figure ~\ref{fig-Example} and can observe two differences. 

\textbf{1. The fine-grained image annotation can partially solve the ``aspect absent'' problem of Multimodal ABSA.} 

According to statistics, there are about 30\% aspects absent in the reviews. For example, in Figure ~\ref{fig-Example}(a), the textual part ``Very comfortable! '' can just convey the ``positive'' opinion, the aspect they are talking about does not appear in the text. In this instance, the fine-grained RoIs (Region of Interest, such as ``bed'') from the related image can be treated as supplements and help to detect the related aspect category ``Room''. 

In contrast, the multimodal ABSA task always uses the whole image representation but ignores exploring the image's fine-grained elements. Moreover, there is no multimodal fine-grained dataset to support this research.

\textbf{2. ``Aspect Category'' in Multimodal ACSA is a good pivot to align the fine-grained elements between image and text. Why not use ``Aspect''?}

We can explore rich fine-grained information from both textual and visual content. How to align them into an overall multimodal representation is the main problem. For example, in Figure ~\ref{fig-Example}(b), the fine-grained elements from the text are ``breakfast'', ``comfortable'' and so on, and the fine-grained elements from images are ``bed'', ``lamp'' and so on. ``Aspect category'' is a general concept, and thus can better align the detailed information between the two modalities. In contrast, ``aspect'' is specific and sometimes absent, so 
we choose the ``aspect category'' to bridge the two modalities.

Based on the above, we can observe that it is necessary to study the multimodal sentiment analysis with the fine-grained exploration both in the text and in the image. Our new constructed dataset MACSA contains more than 21K text-image pairs, and we design six pre-defined categories: \emph{Location}, \emph{Food}, \emph{Room}, \emph{Entertainment}, \emph{Public Area} and \emph{Service}. The MACSA dataset will be described in detail in Section \ref{MACSADataset}.

As mentioned above, for the text with the aspect absent problem, the annotated visual categories can be treated as an important supplement. In order to further study the aspect absent problem, we also build the MACSA-hard dataset as a hard setting that is selected from the MACSA dataset and all textual content in MACSA-hard has aspect absent problem.  
Apart from this, we provide a multimodal graph-based aligned model (MGAM), which jointly models the fine-grained elements from both text and image. Experiments show that our model can facilitate the baseline comparison for future research on this corpus. 
The main contributions of our work are as follows:
\begin{itemize}[leftmargin=*]
\item We propose a Chinese multimodal aspect-category sentiment analysis dataset(MACSA) with multimodal fine-grained aligned annotations. MACSA is the first dataset that contains fine-grained annotations for both image and text. And we also propose a MACSA-hard dataset that has a higher aspect absent ratio to help analyze the aspect absent problem in the Multimodal ACSA task.
\item We design a new multimodal ACSA task to adapt to the multimodal data processing in the real-world situation.
\item We provide a benchmark model MGAM base on multimodal heterogeneous graph. The experimental results show that the MGAM model can facilitate the baseline comparison for future research on these two corpora. 
\end{itemize}

\section{Related Work}
In this section, we briefly review the existing multimodal sentiment analysis datasets and the related works of coarse-grained and multimodal fine-grained sentiment analysis.

\textbf{Multimodal Sentiment Analysis Dataset.}
As the basis of the multimodal sentiment analysis study, the high-quality multimodal dataset is extremely important. In the field of video-based sentiment analysis, there are many publicly available academic datasets~\cite{zadeh2016multimodal,zadeh2018multimodal,busso2008iemocap,poria2018meld,yu2020ch,castro2019towards,li2017cheavd,morency2011towards,perez2013utterance}. But the existing text-image oriented multimodal sentiment analysis datasets are relatively rare, include MVSA~\cite{niu2016sentiment}, Yelp~\cite{truong2019vistanet} and Twitter~\cite{you2016robust}. There are also other multimodal sentiment related datasets, such as Twitter-sarcasm~\cite{cai2019multi} for sarcasm detection task and UR-FUNNY~\cite{hasan2019ur} for humor detection task. Most of them annotated sentiment polarity at the document level or sentence level and were used in coarse-grained sentiment classification tasks. As for multimodal fine-grained sentiment analysis, a Chinese dataset Multi-ZOL~\cite{xu2019multi} and an English dataset Twitter~\cite{yu2019adapting} contained sentiment annotations for reviews at the aspect level. 

However, all the existing corpora for text-image pairs ignore annotating the fine-grained elements in the images. The lack of image fine-grained annotation leads to restrictions on the study of sentiment alignment between the two modalities, which is adverse to studying the multimodal sentiment analysis in more depth.

\textbf{Multimodal Coarse-grained Sentiment Analysis.}
Multimodal coarse-grain sentiment analysis aims to determine the sentiment polarity of a given text-image pair and is the most typical multimodal sentiment analysis task. Many neural network models and multimodal fusion algorithms have been proposed for this task~\cite{borth2013large, wang2014microblog, you2016robust, cao2016cross, you2016building, xu2018co, truong2019vistanet, cai2019multi, xu2020reasoning}. Most of the works have obtained a similar conclusion, that is, the images play a supporting role to text, which is the same as we observed from the fine-grained annotations of the corpus. Besides, we find that images also play the role of supplementing semantics and eliminating ambiguity. The information expressed by images is more objective than the text. It is necessary to capture a certain amount of common senses and knowledge to understand the sentiment expressed by images.

\begin{figure*}[htp]
    \centering
    \includegraphics[width=14cm]{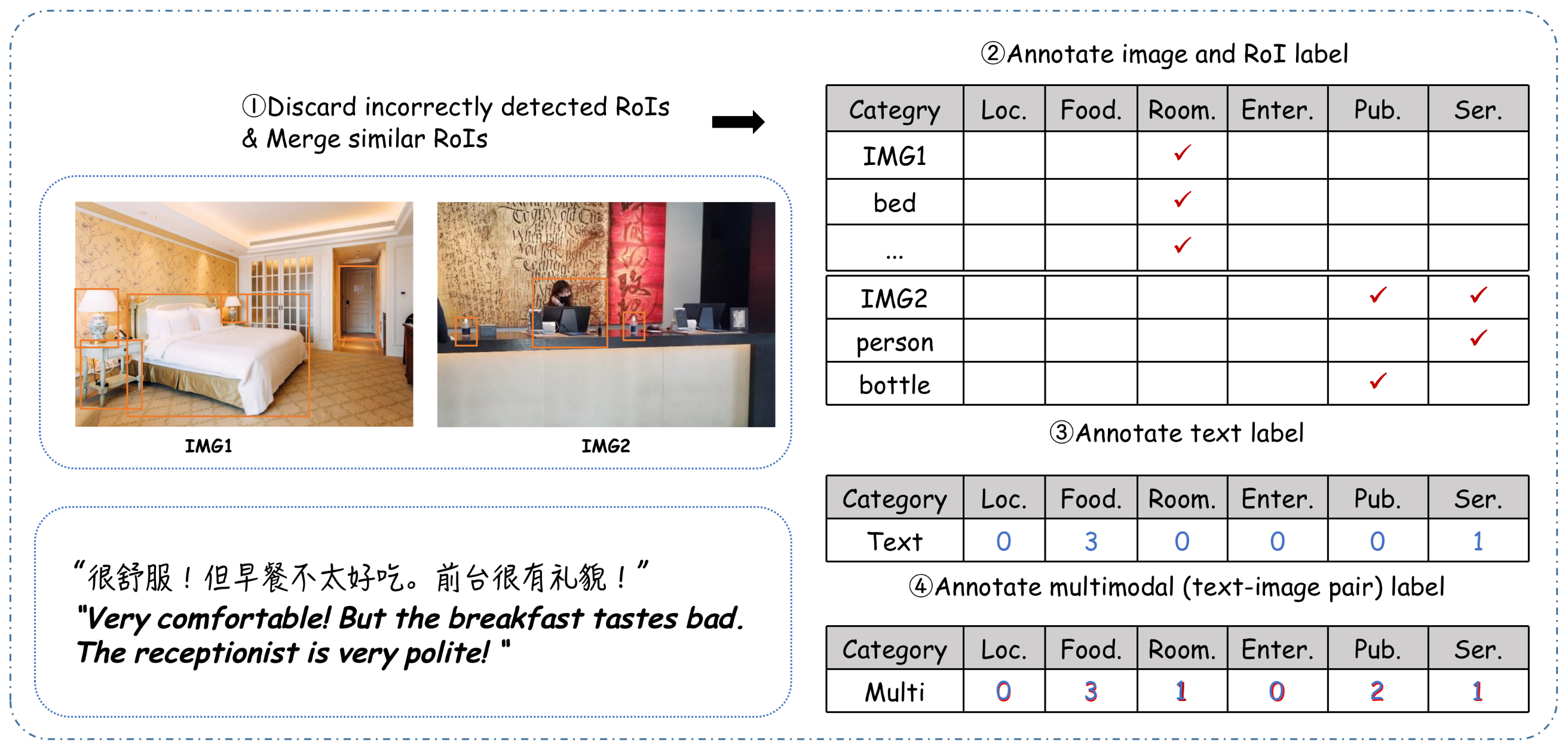}
    \caption{MACSA dataset annotation. The left part shows the input text and images, and the framed areas on the image are the recognized RoIs. The table on the right shows the fine-grained annotation of unimodal and multimodal data. For the image, the annotator annotates the images and the correctly detected RoIs and uses the ``\checkmark '' to represent that the image or RoI is related to the aspect category. Otherwise, it is irrelevant. For the text, we use  $\lbrace 0, 1, 2, 3 \rbrace$ to represent the label  ``irrelevant'', ``positive'', ``neutral'' and ``negative'' . The multimodal labels are the same as the text labels, but the annotator needs to consider the information of text and image at the same time.}
    \label{fig:ann}
\end{figure*}

\textbf{Multimodal Fine-grained Sentiment Analysis.}
Early works on fine-grained sentiment analysis only focused on text-level~\cite{wang2016attention, xue2018aspect, hu2018can, zhu2019aspect, li2020sentence}. But for multimodal data, the goal becomes to identify fine-grained sentiment elements in multimodal text-image pairs. In 2019, \citet{xu2019multi} proposed aspect based multimodal sentiment analysis task and proposed a novel Multi-Interactive Memory Network (MIMN) model based on BiLSTM for this task. \citet{yu2019adapting} proposed a BERT-based multimodal architecture TomBERT for target-oriented multimodal sentiment classification task (TMSC). \citet{yu2019entity} proposed a entity-sensitive attention and fusion network for multimodal target-based sentiment classification. \citet{ju2021joint} proposed a multimodal joint learning approach with auxiliary cross-modal relation detection for multimodal aspect-level sentiment analysis. \citet{khan2021exploiting} used an object-aware transformer to translate images to language for cross-modal fusion. The datasets of the existing work tend to focus on the fine-grained annotation of text, and the image is in a subordinate position. This leads to some models after adding visual-modal information is not significant compared with the text-based model. And most of the existing datasets only contain one picture, which is not enough to fit the actual situation. To mitigate this, we propose a new multimodal fine-grained dataset and task. We will introduce them in detail in the following section.

\section{MACSA Dataset}\label{MACSADataset}
In this section, we define the multimodal ACSA task and introduce a novel Chinese multimodal aspect-category sentiment analysis dataset with multimodal fine-grained aligned annotations.

\subsection{Task Definition} 
As shown in Figure ~\ref{fig-Example}(b), for each given text-image pair, there is a textual content $T= \lbrace W_{1}$, ... , $W_{L}\rbrace$ and equipped several images $I=\lbrace I_{1}$, ... , $I_{K}\rbrace$. We pre-defined an aspect category set $A=\lbrace A^{1}$, ... , $A^{N} \rbrace$. Here, $L$, $K$, and $N$ respectively represent the number of words in the sentence, the number of images, and the number of pre-defined aspect categories. The goal is to learn a multimodal sentiment classifier at the aspect category level that can correctly detect the related categories of multimodal data, and predict the sentiment labels(``irrelevant'', ``positive'', ``neutral'' and ``negative'') for the related categories in unseen samples.

\subsection{Data Collection}
 To ensure the authenticity and reliability of both the text and image, we collected 120K user-generated original multimodal reviews from a large Chinese online travel website ``www.qunar.com'' which provides hotel online booking services. And then, we selected the reviews with up to six pictures, and the max review's length $L$ is 512. Finally, we collected 24K reviews with images as the unlabeled data.
 
 Because we need to annotate the fine-grained elements for image modality, to facilitate the image annotation,
 we used a object detection tool detectron2~\cite{wu2019detectron2} to automatically detect the region of interesting (RoI) from images. Finally, through multimodal fine-grained annotation and dropping the data with serious conflicts of sentiment annotation, we obtained 21K text-image pairs with RoIs as our MACSA dataset. The data in the MACSA dataset are all public comments of users and do not involve personal privacy.

\subsection{Data Annotation}
To ensure the independence of different modality labels, we invited twelve professional annotators to participate in data annotation. We will introduce the annotation in detail.

\textbf{Aspect Category Definition.} Firstly, in order to scientifically divide the aspect category of the hotel-domain data, we invited a professional doctorate in management and two experienced researchers who work on Natural Language Processing (NLP) to define six aspect categories of the data. Through statistical analysis of the original data, we select ten categories utilizing word frequency and LDA (Latent Dirichlet Allocation)~\cite{blei2003latent}. Then, by combining with the management knowledge of the hotel, we manually screened out six categories that can be judged independently from image or text, and cover the attention of users in the hotel field as far as possible, forming our aspect category criterion: \emph{location}, \emph{food}, \emph{room}, \emph{entertainment}, \emph{service} and \emph{public area.}  

\textbf{Annotation.} 
For each text-image pair, we need to  annotate the following elements:
\begin{itemize}[leftmargin=*]
    \item \textbf{Text Annotation}: refers to annotating the related categories and sentiments according to the textual content.
    \item \textbf{Image Annotation}: refers to annotating the related categories in each image and RoI. 
    \item \textbf{Text-Image Pair Annotation}: refers to annotating the related categories and sentiments according to the whole pair.
\end{itemize}

What are the fine-grained elements in image modality, and how to align the fine-grained elements obtained from the two modalities are the most important problems. 

For each image, the RoIs in the image are fine-grained elements with rich information. Thus we first recognize RoIs using detectron2 and then do the filtering and merging pre-processing.
Specifically, annotators are asked to identify the correct RoIs and merge similar RoIs. The wrong detect RoIs and redundant RoIs are removed from the data.

We propose to use the aspect category as a pivot that aims to align the fine-grained elements from multimodal inputs. According to the same aspect category criterion, we first annotate the relevance of the text modality, image modality, and text-image pair to the six pre-defined categories with the labels 0 (irrelevant) or 1 (relevant). 
And then, for the text modality and text-image pair, we manually recognize the sentiments to the related categories with the sentiment states 1 (positive), 2 (neutral) and 3 (negative).

It is noteworthy that we annotate the relevance of images and RoIs to pre-defined aspect categories. And since the visual content seldom conveys the sentiments, we did not annotate the sentiments for each image and RoI. 
We provide a detailed example of data annotation in Figure ~\ref{fig:ann} as a reference.
In the labeling stage, the text modality, image modality, and multimodal data were annotated under the same aspect category criterion. And the annotator can only see the current modality when annotating. 

To ensure the quality of annotation, each text-image pair is annotated by five annotators. The final label is determined by the label with the highest proportion and we dropped the data with serious conflicts of sentiment annotation.
In order to verify the annotation quality, we calculated the kappa coefficient of the annotation consistency; the kappa coefficients are 0.78, 0.83, 0.89, and 0.74 for the text, image, RoI, and text-image pair data.

By randomly sampling the annotated data, we divided the multimodal data into the training set, validation set, and test set. The statistics of the MACSA dataset are reported in Table \ref{data}.

\subsection{Data Analysis}
After the data annotation, we conducted a statistical analysis of the dataset in Table ~\ref{data}. Compared to other multimodal fine-grained sentiment analysis datasets, the special feature of the MACSA dataset is that the dataset contains the fine-grained annotations for images and contains multimodal fine-grained alignment annotations at the aspect category level, while the previous multimodal sentiment analysis dataset only contains the fine-grained annotation for textual content.


\textbf{Multimodal fine-grained aligned data is helpful for aspect absent problem. } 
We found that there are 28.60\% more sentiment labels in multimodal data than in only-text-modality data. This means that multimodal data contains more abundant sentiment information, and mining the sentiment clues needs to comprehensively consider the information of two modalities. Furthermore, some difficult problems appearing under text modality may be easily solved by using multimodal data, such as the aspect absent problem in text-based ACSA. Since the aspect does not appear in the text, it's difficult to recognize the aspect category just from the single text.
As shown in Figure ~\ref{fig-Example}, in the textual content``Very comfortable! '', the described aspect category ``Room.'' is absent. But we can recognize the ``bed'' RoI from visual content, which indicates the ``Room.'' category.
Our dataset can provide plentiful multimodal clues which are helpful to partially solve the aspect absent problem.

\begin{table}

\centering
\caption{
Statistics of the MACSA dataset and the MACSA-hard dataset. The datasets are randomly divided into training set, validation set and test set in a ratio of 10:1:1.
}
\begin{tabular}{ccrrr}
\hline
\multicolumn{2}{c}{Dataset}             & Review & Image &RoI \\ 
\hline
                   & Train.      & 17,508   & 13,131 & 29,794              \\ 
        {MACSA}             & Valid.     & 1,800    & 1,344  & 3,012               \\
                            & Test.       & 1,800    & 1,350  & 3,197               \\ \cline{2-5} 
                            & Total.      & 21,108   & 15,825 & 35,997              \\ \hline
& Train.      & 4,260    & 3,219  & 7,324               \\ 
        {MACSA-hard}             & Valid.     & 396     & 286   & 598                \\  
                      & Test.       & 402     & 302   & 701                \\ \cline{2-5} 
                            & Total.      & 5,058    & 3,807  & 8,623               \\ \hline
\end{tabular}

\label{data}
\end{table}

To further study the aspect absent problem in the multimodal ACSA task, we selected the data that there is at least two aspect category absent in textual data to construct the MACSA-hard dataset. This dataset could be used as a platform for the researchers to analyze the aspect absent problem. The statistics of the dataset are reported in Table \ref{data}.

\begin{figure}[ht]
    \centering
    \includegraphics[width=8cm]{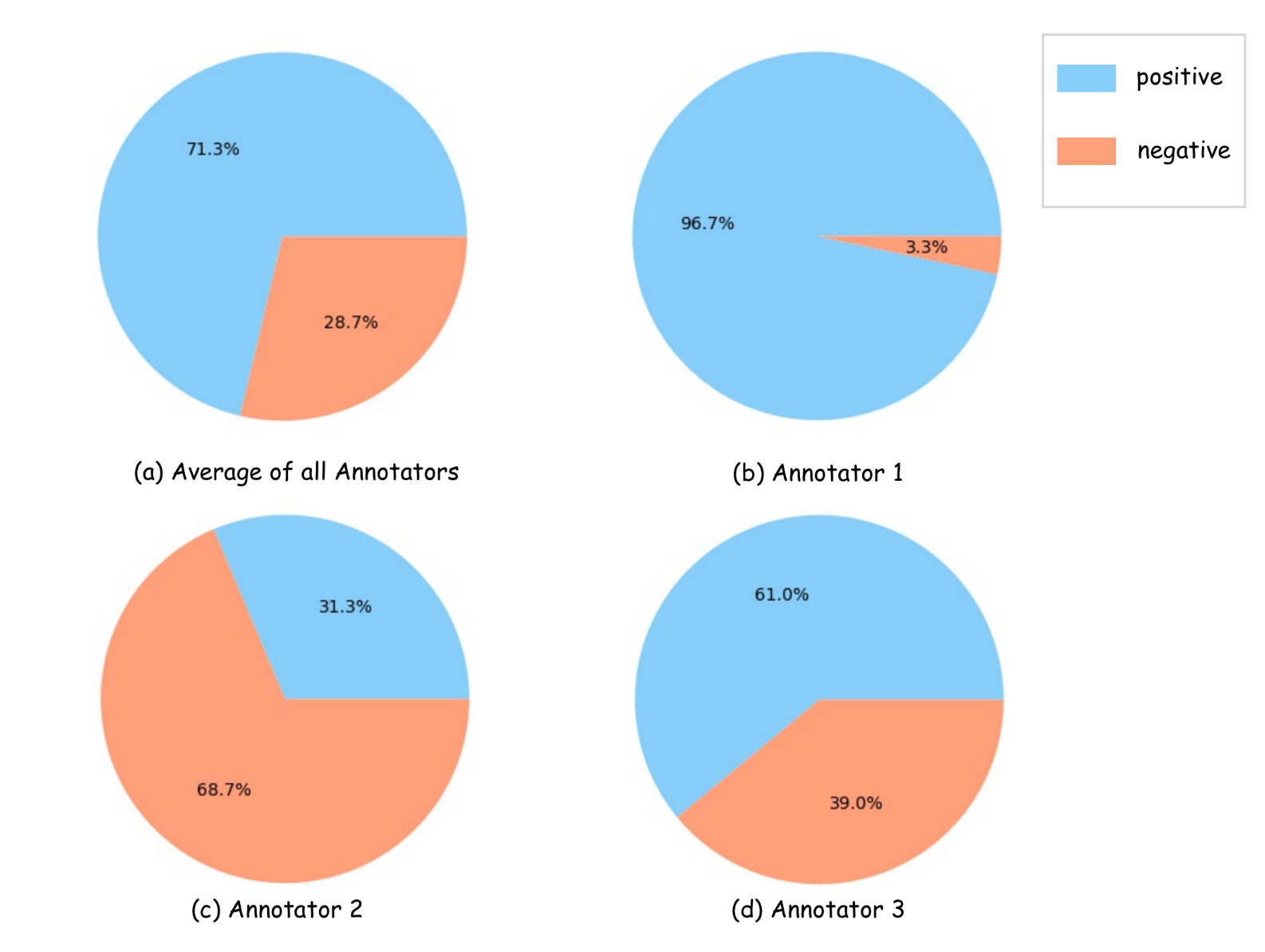}
    \caption{The ratio of sentiment annotations of the image by different annotators.}
    \label{image_s}
\end{figure}

\textbf{The sentiment of visual content in the MACSA dataset is equivocal. }
In addition to focusing on the semantic complement information provided by fine-grained elements in images, we also explored visual sentiment in our dataset. We tried to manually annotate the visual content in the dataset with sentiment labels and reported the average of all annotators' and partial annotators' sentiment labeling proportion in Figure ~\ref{image_s}. The results showed that annotators are inconsistent in labeling most of the data, the Kappa coefficient of the visual sentiment annotation was only 0.23. Based on our observation, we regard that the lack of expressions and actions of the characters in the visual content makes it difficult to determine the sentiment polarity of images. Therefore, we assumed that the sentiment of non-portrait images often is only understood if combined with the textual context.

\textbf{MACSA dataset is suitable for other multimodal tasks. }
The multimodal ACSA task is just one typical task that the MACSA dataset is suitable.
Furthermore, the MACSA dataset can be widely applied to other tasks, for example, the text-based fine-grained sentiment analysis and other multimodal tasks, such as image-to-opinion generation and so on.
We will explore more tasks and applications of the MACSA dataset in the future. For example, we tend to add sentiment cause annotations to make the dataset suit the multimodal sentiment cause discovery task in the next release.

\section{Methodology}
\begin{figure*}[htp]
    \centering
    
    \includegraphics[width=14cm]{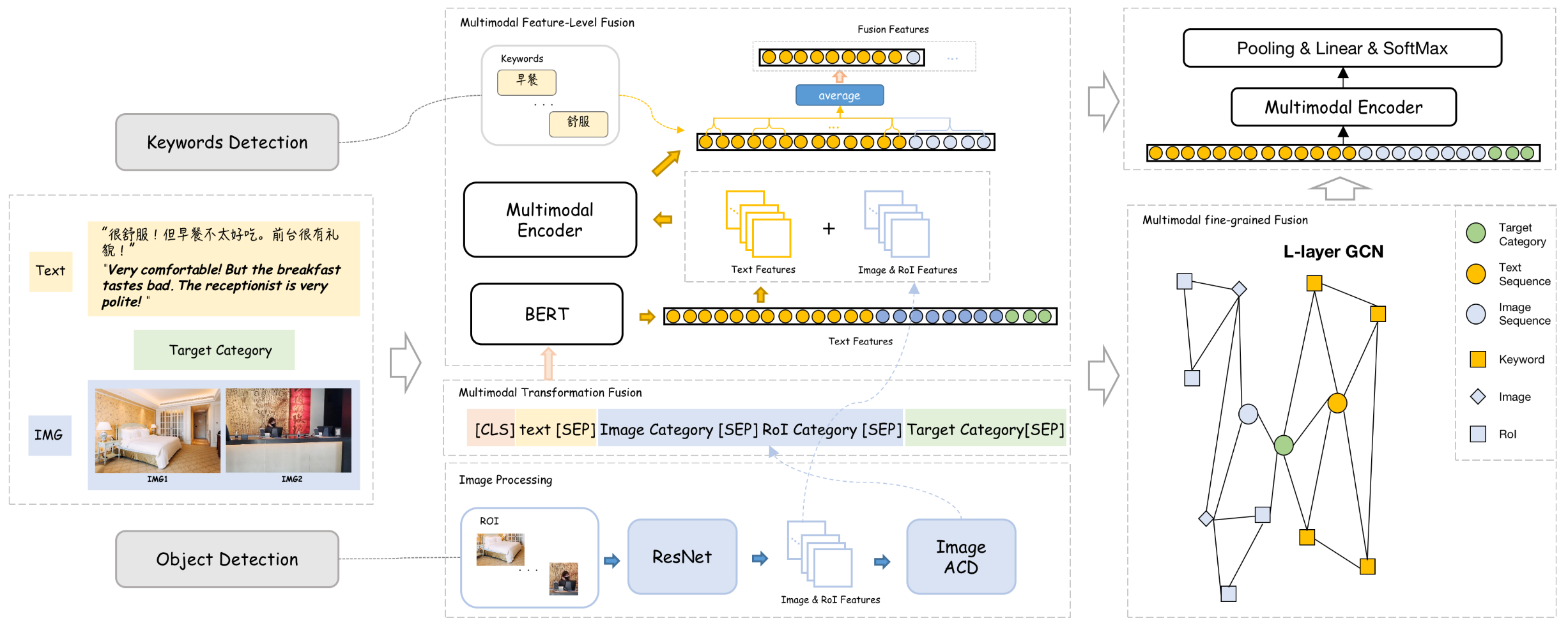}
    
    \caption{Overview of MGAM model for multimodal ACSA.}
    \label{fig:model}
\end{figure*}

According to the characteristics of the dataset that aligns two modalities with the category as a bridge, we adopted the heterogeneous graph to build the relationship of fine-grained elements. And we proposed the multimodal graph-based ACSA model (MGAM) using convolutional graph neural network(GCN)\cite{kipf2016semi}, as illustrated in Figure ~\ref{fig:model}. In this section, we describe the MGAM in detail. The framework of the model includes three modules: an image processing module, a feature extraction module, and a heterogeneous graph-based multimodal fine-grained fusion module. 

In the following subsections, we will introduce each module in detail.

\subsection{Image Processing}
The image processing module includes two tasks: extracting fine-grained elements in the image, and aspect category detection of the images and the extracted visual elements. We define the input images set as $I=\lbrace I_{1}$, ... , $I_{K}\rbrace$. By using Detectron2 on each image $I_{k}$, we obtain a RoI set $R_{k}=\lbrace R_{1},...,R_{J_{k}} \rbrace$ , where $J_{k}$ denotes the number of RoIs. Combine the obtained $R_{k}$ to obtain a unified RoI set $R=\lbrace R_{1},...,R_{J} \rbrace$ , where $J$ denotes the number of RoIs. Then we use the image as input to train a multi-label image classification model to predict the related categories to the image as the input of the second stage.
Following~\citet{krizhevsky2012imagenet} and \citet{he2016deep}'s work, we use the CNN-based models for this task. 

Similar to the image, the RoI identified from the image uses the same processing method, and finally, we get the related category set of the images $C^{I}=\lbrace C^{I}_{1}$, ... , $C^{I}_{N}\rbrace$ and the related category set of the RoIs $C^{R}=\lbrace C^{R}_{1}$, ... , $C^{R}_{N}\rbrace$, where $C^{I}_{n}, C^{R}_{n} \in \lbrace 0, 1 \rbrace$ represent whether the images set and the RoIs set are related to the pre-defined category $A^{n}$.

\begin{gather}
R_{k} = Detectron2(I_{k}), k\in [1,K] \tag{1}\\
C^{I} = CNN(I), I=\lbrace I_{1}, ... , I_{K} \rbrace \tag{2}\\
C^{R} = CNN(R), R=\lbrace R_{1},...,R_{J} \rbrace \tag{3}
\end{gather}

\subsection{Feature Extraction}\label{feature-extraction}

\textbf{Visual Feature Extraction.}
We adopt one of the state-of-the-art image recognition models called ResNet-152 (res5c)\cite{he2016deep} to extract visual features of the images and RoIs. For a given input image, we first resize it to 224×224 pixels, and then use the ResNet model to obtain the last layer representation $ResNet(I_{k})$, and use a linear function to transform the visual features to the same space of textual features as in \cite{yu2019adapting}, where $W_{v}\in \mathbb{R}^{d\times 2048}$ is the learnable parameter.
\begin{gather}
ResNet(I_{k}) = \lbrace r_{j}|r_{j}\in \mathbb{R}^{2048}, j=1,2,...,49\rbrace \tag{4}\\
v^{I_{k}} = W_{v}ResNet(I_{k}) \tag{5}\\
v^{R_{j}} = W_{v}ResNet(R_{j}) \tag{6}
\end{gather}

After getting the feature of each image, we concatenate all of the images’ features as a unified image feature. The RoI’s feature is as same as the image. 
\begin{gather}
v^{I} = \lbrace v^{I_{k}}|v^{I_{k}}\in \mathbb{R}^{d\times 49}, k\in [1,K] \rbrace \tag{7}\\
v^{R} = \lbrace v^{R_{j}}|v^{R_{j}}\in \mathbb{R}^{d\times 49}, j\in [1,J] \rbrace \tag{8}
\end{gather}

For each image or RoI, we concatenate the image feature $v^{I_{k}}$ or the RoI feature $v^{R_{j}}$ with the text feature $v^{T}$ together and feed them to multimodal encoder ($MEncoder$) as in \citet{yu2019adapting} to automatically model the interactions between textual and visual features.
\begin{gather}
v^{T\&I} = concat([v^{T},v^{I_{k}}]) \tag{9}\\
H^{T\&I} = MEncoder(v^{T\&I}) \tag{10}
\end{gather}
The $H^{T\&I}=\lbrace h^{T_{0}},...,h^{T_{L}}, h^{I_{0}} ,...,h^{I_{49}}\rbrace$, where the L represent the length of text. We average the image/RoI tokens in the final hidden state to produce the image/RoI representation $h^{I}\in \mathbb{R}^{d \times 768}$, where K is the number of images.
\begin{gather}
h^{I} = Avg(h^{I_{0}}, ...,h^{I_{49}})\tag{11}\\
H^{I} = \lbrace h^{I_{0}}, h^{I_{1}}, ...,h^{I_{K}}\rbrace \tag{12}
\end{gather}

\textbf{Textual Feature Extraction.}
Due to the promising performance of BERT model in the field of natural language processing, we adopt Chinese-based BERT to extract the textual features. we tried to transform each sentence S into four sub-sentences: the target aspect category $A^{T}=( W_{1}^{T},W_{2}^{T}$,...,$W_{S}^{T})$, the text content $T=( W_{1},W_{2}$,...,$W_{L})$, the  images' related category $A^{I}=(W_{1}^{I},W_{2}^{I}$,...,$W_{S_{I}}^{I})$ and the RoIs' related category $A^{R}=( W_{1}^{R},W_{2}^{R}$,...,$W_{S_{R}}^{R})$, where $S, L, S_{I}, S_{R}$ are the length of the text. And the final input sentence could be expressed as: ``[CLS]$T$[SEP] $A^{I}$[SEP] $A^{R}$[SEP]$A^{T}$[SEP]''. Then we feed the input sentence to the BERT embedding layer to obtain the textual embedding sequence $x=( x_{0},x_{1},...,x_{L+S_{I}+S_{R}+S+5} )$, and then send it into BERT encoder to obtain the final hidden state of the first [CLS] token:
\begin{gather}
h_{[CLS]} = BERT-Encoder(x) \tag{13}\label{fusion-tag}
\end{gather}

\subsection{Multimodal Fusion Module}
Our multimodal fusion module simultaneously utilizes two commonly used multimodal fusion methods: multimodal feature-level fusion and multimodal transformation fusion. And we also propose a multimodal heterogeneous graph-based fine-grained fusion method, which can align the fine-grained information of the multimodal data. We will describe each part of the multimodal fusion module in detail.

\textbf{Multimodal Fine-grained Fusion.}
We obtain fine-grained elements of the visual modality from the image processing stage. Meanwhile, we use a Chinese analysis tool Jieba\footnote{https://github.com/fxsjy/jieba} to obtain the keywords(nouns, adjectives and adverbs) of the text as fine-grained elements of the textual modal.

As fine-grained elements from text and images are in different feature spaces, we choose the target category as the intermediate node to align the interrelated fine-grained elements and establish the multimodal fine-grained heterogeneous graph to build cross-modal relationships. We use the features which are extracted in section \ref{feature-extraction} to initialize these fine-grained element nodes. Because the tokenization of the Bert model for Chinese is character-level, in order to obtain the keywords’ representation containing contextual information, we average the representation of each character in the keyword as the word-level representation.

As shown in Figure ~\ref{fig:model}, the multimodal heterogeneous graph is composed of textual nodes and visual nodes. Textual nodes include the target category node, the text sequence's nodes, and the keywords' nodes. Visual nodes include the image sequence nodes, the images' nodes, and the RoIs' nodes. We concatenate the feature of these nodes to produce hidden state vectors $G = \lbrace g_{0},g_{1},g_{2},...,g_{T} \rbrace$, where $g_{t}\in \mathbb{R}^{d \times 768}$ represents the hidden state vector of the nodes and $T$ represents the number of nodes.
Inter-text nodes use the attention mechanism to score the correlation and construct edges for nodes whose correlation exceeds the predetermined threshold (0.25). There is a fixed edge between the target category node and the text sequence node. The image node contains the direct edge between the image node and the dependent RoI node. There are fixed edges between image nodes and the image sequence node. 
After establishing the multimodal heterogeneous graph, we first project the multimodal nodes onto a unified feature space followed by a non-linear activation function $tanh$. And we generate the feature of nodes $H = \lbrace h_{0},h_{1},h_{2},...,h_{T} \rbrace$, where $h_{t}\in \mathbb{R}^{d \times 768}$represents the feature vector of the nodes. Thereafter, we use the $L$-layer GCN\cite{zhang-etal-2019-aspect} model to compute the graphical representation of the target category on the multimodal heterogeneous graph. We update the representation of each node with graph convolution operation with normalization factor\cite{zhang2018graph} as below:
\begin{gather}
h_{i} = tanh(W^{h}g_{i}+b^{h})\tag{14}\\
\Tilde{h_{i}^{l}} =  \sum_{j=1}^{n} A_{ij}W^{l}h_{j}^{l-1}\tag{15}\\
h_{i}^{l} = RELU(\Tilde{h_{i}^{l}}/(d_{i}+1)+b^{l}) \tag{16}\\
V^{G} = h_{0}^{L} \tag{17}
\end{gather}
where $h_{j}^{l-1}\in \mathbb{R}^{d \times 768}$  is the $j$-th token’s representation evolved from the preceding GCN layer while $h_{i}^{l}\in \mathbb{R}^{d \times 768}$ is the product of current GCN layer, and $d_{i}=\sum_{j=1}^{n} A_{ij}$ is degree of the $i$-th token in the graph. The weights $W^{h}$, $W^{l}$ and bias $b^{h}$, $b^{l}$ are trainable parameters. Finally, we obtain the multimodal heterogeneous graph based fine-grained fusion feature $V^{G}\in \mathbb{R}^{d \times 768}$.

\textbf{Multimodal Transformation Fusion.}
The multimodal transformation fusion is to select a target modality, then transform another modality into the form of the target modality and fuse them in the same feature space so that the representation of the target modality contains the semantic information of another modality. In our fusion module, we use the textual content as the target and transform the other modality as the text of the related category. We obtain the multimodal transformation fusion features in Equation \ref{fusion-tag}.

\textbf{Multimodal Feature-level Fusion.}
The multimodal feature-level fusion is to fuse each unimodal representation at the feature level to obtain a multimodal representation containing the semantic information of each modality. 

We adopt the feature-level fusion method on visual feature extraction section. And in the final stage, we concatenate the graphical representation of the target category $V^{G}\in \mathbb{R}^{d \times 768}$, the image features $V^{I}\in \mathbb{R}^{d \times 768\times K}$, the RoI features $V^{R} \in \mathbb{R}^{d \times 768\times J}$ and text features $h_{[CLS]}$ 
together, and feed them to the another multimodal encoder($MEncoder$) to learn the interactions between each features. 

\begin{gather}
v^{M} = concat([h_{[CLS]},h^{I},h^{R},v^{G}]) \tag{18}\\
H = MEncoder(v^{M}) \tag{19}
\end{gather}

Finally, we obtain the final hidden state of the ``[CLS]'' token $H$, which is used for sentiment classification.

\begin{table*}[]
\centering
\caption{ 
Experiment results on MACSA dataset and MACSA-hard dataset for multimodal ACSA task. The BiLSTM-text model and BiLSTM-pair model are weakened versions of MACSA-BiLSTM, only using text input and text-image pair input. And the MGAM-w/o graph model does not use the fine-grained features based on the multimodal heterogeneous graph. The models with $^{*}$ marks are modified in the image embedding part to fit the multi-image task settings.
}
\begin{tabular}{cccccccccc}
\hline
Modality   & Method & \multicolumn{4}{c}{MACSA}                                                                & \multicolumn{4}{c}{MACSA-hard}                                                           \\ \cline{3-10} 
                             &                         & Acc.           & Pre.                       & Rec.                       & F1        & Acc.           & Pre.                       & Rec.                       & F1        \\ \hline

Text        & BiLSTM-text            & 63.72          & \multicolumn{1}{l}{45.64} & \multicolumn{1}{l}{39.79} & 41.00          & 46.51          & 31.86                      & 33.35                      & 32.38          \\ 
                             & SynATT                  & 76.50           & \multicolumn{1}{l}{59.57} & \multicolumn{1}{l}{61.29} & 57.10          & 46.52          & \multicolumn{1}{l}{23.52} & \multicolumn{1}{l}{29.20}  & 24.85          \\ 
                             & GCAE                    & 76.44          & \multicolumn{1}{l}{65.51} & \multicolumn{1}{l}{57.88} & 57.61          & 47.76          & \multicolumn{1}{l}{25.37} & \multicolumn{1}{l}{29.75} & 24.02          \\ \hline
Text\&Image & BiLSTM-pair     & 72.05          & 60.70                       & 51.88                      & 55.16          & 52.99          & 36.40                      & 34.92                      & 35.28          \\ 
                             & MIMN$^{*}$                    & 72.94          & 64.36                      & 53.13                      & 57.05          & 51.74          & 38.18                      & 37.60                      & 36.92          \\ 
                             & MACSA-BiLSTM            & 73.00          & 58.14                      & 59.83                      & 58.57          & 53.48         & 39.97           & 39.94          & 38.15                \\ \hline

Text                         & BERT                    & 80.25          & 73.42                      & 69.29                      & 70.96          & 71.67          & 62.26                      & 60.70                      & 60.78          \\ \hline
Text\&Image & TomBERT$^{*}$                 & 80.84          & \multicolumn{1}{l}{74.56} & \multicolumn{1}{l}{70.59} & 71.96           & 71.89          & 61.90                      & \textbf{63.20}                      & 62.07          \\ 
                             & MGAM-w/o graph              & 84.12 & 76.04             & 75.82             & 75.74 & 73.94 & 64.87             & 62.11             & 62.95 \\ 
                             & MGAM             & \textbf{86.06} & \textbf{78.31}             & \textbf{76.01}             & \textbf{77.68} & \textbf{75.25} & \textbf{72.41}             & 60.89             & \textbf{64.17} \\  \hline
\end{tabular}

\label{result}
\end{table*}
\section{Experiments}
In this section, in order to evaluate our dataset and model, we compared our model with existing models on the MACSA and MACSA-hard datasets.

\subsection{Baseline}
In this part, We compare our model with a series of baseline methods on the MACSA dataset. And due to some of these models that we marked $^{*}$ are originally designed for single-image tasks. In order to transfer the model to the Multimodal ACSA task for comparison, we randomly choose one image from related images. In addition, we didn't compare our models with existing multimodal pre-training models\cite{radford2021learning,qi2020imagebert,su2019vl,tan2019lxmert} as the language used in our datasets and the visual setting are different from the pre-training modals' corpus. 

\begin{itemize}[leftmargin=*]

\item\textbf{GCAE:} The Gated Convolutional network with Aspect Embedding (GCAE)~\cite{xue2018aspect} based on convolutional neural networks and gating mechanisms. Use the Tanh-ReLU gate unit to select the output sentiment features for a given aspect category.
\item\textbf{LSTM+SynATT:} The LSTM+SynATT model ~\cite{he2018effective} imposed some syntactical constraints on attention weights for the textual modality.
\item\textbf{MIMN$^{*}$:} The Multi-Interactive Memory Network ~\cite{xu2019multi} learn the interactive influences in cross-modality and self-modality.
\item\textbf{MACSA-BiLSTM:}  As the BiLSTM model could effectively model sequence content. We propose the MACSA-BiLSTM model, which models the text content, images, RoIs, and the category detection result as a multimodal sequence and adopts the Multi-Interactive Memory Network to learn the fusion representation for the Multimodal ACSA task.
\item\textbf{BERT:} The BERT model~\cite{devlin2018bert} is the representative model of the large pre-trained language model, which has strong text representation ability and can learn alignment between two arbitrary inputs.
\item\textbf{TomBERT$^{*}$:} The TomBERT~\cite{yu2019adapting} models the inter-modal interactions between visual and textual representations. And this model adopts a Target-Image (TI) matching layer to obtain a target-sensitive visual representation.
\end{itemize}

\subsection{Implementation Details}
We build our BERT-based models on top of the pre-trained uncased Chinese BERT model released by~\citet{devlin2018bert} and tune the hyperparameters on the development set of the MACSA dataset. We set the learning rate as 5e-5, the number of attention heads as m = 12, and the dropout rate as 0.1. All the bert-based models are fine-tuned for eight epochs. We set the GCN layer number L as 3, and the maximum number of nodes in a multimodal heterogeneous graph is 20. In the image processing section, as the number of images has been restricted during the dataset collection, the maximum number of pictures is six as same as the max number of objects.
AS for the BiLSTM-based model, we use Jieba Chinese word segmentation tool for word segmentation. Then we use SGNS~\cite{li2018analogical} to initialize all word embeddings to 300-dimensional vectors (pre-trained on Baidu Encyclopedia corpus). For a fair comparison, in each experiment, we select five same random seeds and report the average performance of five times. And all the models are implemented based on PyTorch with two NVIDIA TeslaV100 GPUs.

\subsection{Experimental Results and Analysis}
We adopt accuracy and macro-F1 scores to compare the performances of the models. Experimental results are reported in Table \ref{result}. According to the experimental results, we get the following conclusions.

Firstly, our MGAM-w/o graph model, which does not contain fine-grained features based on the multimodal heterogeneous graph, achieves impressive F-score results than text-based BERT that uses text-only data on the MACSA dataset. And also performs better than $TomBERT^{*}$ model which only contains one image as visual input. This verifies that images provide additional information, which is helpful for the multimodal ACSA task.

Secondly, our MACSA-BiLSTM model and MGAM-w/o graph model, which add fine-grained images and RoIs annotation, outperform the other similar structured baselines(BiLSTM-pair and TomBERT$^{*}$), showing the potential usage of fine-grained aligned information between images and text for the multimodal ACSA task. Our two baseline methods can better integrate these multimodal fine-grained elements also demonstrate that it is effective in using the two multimodal fusion methods: multimodal feature-level
fusion and multimodal transformation fusion simultaneously. 

Thirdly, on the MACSA-hard dataset, our MGAM model achieves competitive results while other methods perform poorly. This verifies that the multimodal fine-grained information, especially the elements from the images, can help to alleviate the aspect absent problem. And it shows that the multimodal aligned information is helpful to the multimodal aspect category detection task. This significantly helps the model achieve better performance in the category-based sentiment analysis task.

\begin{figure*}[ht]
    \centering
    
    \includegraphics[width=14cm]{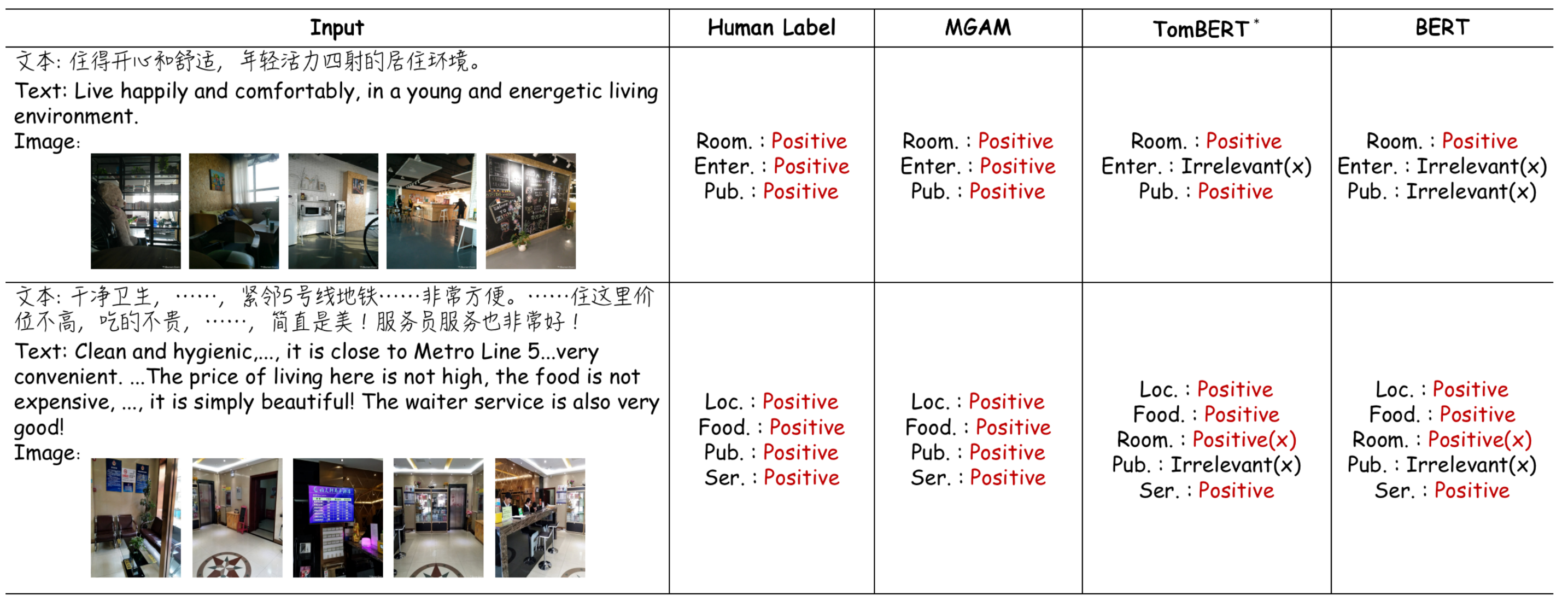}
    
    \caption{Case analysis on MGAM, TomBERT$^{*}$, BERT. The cross label means different prediction than human label.}
    \label{fig:case}
\end{figure*}

Fourthly, our MGAM model achieves better performance than the MGAM-w/o graph model. This observation is in line with our intuition that the multimodal heterogeneous graph is able to align the useful fine-grained element for the target aspect category from multimodal data and is helpful for multimodal fusion and modeling complementary information between multimodal data. 

Finally, the MGAM model performs best, indicating that the fine-grained alignment information between images and texts could be used on BERT to obtain additional improvements.


\textbf{Image/RoI Category Detection Result.} For unlabeled image data, to obtain the predicted label of the image or RoI under our category criterion, we trained two image category detection algorithms based on CNN~\cite{krizhevsky2012imagenet} and VGG~\cite{simonyan2014very} (which adopts a smaller convolution kernel and more convolution sublayers) models on MACSA dataset. The experimental results show that VGG (92.92\% in Accuracy) performs better than the common CNN model (77.61\% in Accuracy), thus we use the VGG model for the image/RoI category detection task. 



\subsection{Ablation Experiment}
To analyze the effectiveness of each module in the MGAM method, we carried out a series of ablation experiments on the MACSA dataset. From Table~\ref{abla}, we can find that the fine-grained fusion module improves the model's ability to capture the sentiment cues in multimodal data. By ablating the nodes of different modalities in the multimodal heterogeneous graph, we explore the performance of the model under the setting of the full graph structure, the single-modal graph structure, and the setting of discarding the entire heterogeneous graph which are denoted as ``-w/o Textual Nodes'', ``-w/o Visual Nodes'' and ``-w/o Graph Nodes'' in Table~\ref{abla}.

From these results, we can observe that:(1)the proposed MGAM consisting of all modules achieves the best performance on the MACSA dataset; (2)removing the textual nodes or the visual nodes in the graph drops performance, but not drastically, which demonstrates the effectiveness of intra-modal fine-grained fusion; (3) removing all the graph nodes drops performance by about 2\%, verifying the complementary role of fine-grained elements in different modalities. The models without a completed graph structure perform worse, indicating that the fine-grained fusion method based on the multimodal heterogeneous graph is beneficial to improving the performance of the Multimodal ACSA task. 

\begin{table}[]

\caption{\label{citation-guide}
 We ablate the fine-grained element nodes(Textual Nodes and Visual Nodes) in the multimodal heterogeneous graph to validate the effect of each part.
}

\begin{tabular}{cccccll}
\cline{1-5}
Method             & \multicolumn{4}{c}{MACSA}                                                                                                       &  &  \\ \cline{2-5}
                                    & \multicolumn{1}{c}{Acc.}           & \multicolumn{1}{c}{Pre.}           & \multicolumn{1}{c}{Rec.}           & F1.            &  &  \\ \cline{1-5}
MGAM                        & \multicolumn{1}{c}{\textbf{86.06}} & \multicolumn{1}{c}{\textbf{78.31}} & \multicolumn{1}{c}{\textbf{76.01}} & \textbf{77.68} &  &  \\ 
\multicolumn{1}{l}{-w/o Textual Node}                                & \multicolumn{1}{c}{85.92}          & \multicolumn{1}{c}{79.75}          & \multicolumn{1}{c}{75.35}          & 76.92          &  &  \\ 
\multicolumn{1}{l}{-w/o Visual Node} & \multicolumn{1}{c}{85.33}          & \multicolumn{1}{c}{78.72}          & \multicolumn{1}{c}{75.85}          & 76.90          &  &  \\ 
\multicolumn{1}{l}{-w/o Graph Node}      & \multicolumn{1}{c}{84.12}          & \multicolumn{1}{c}{76.04}          & \multicolumn{1}{c}{75.82}          & 75.74          &  &  \\ \cline{1-5}
\end{tabular}

\label{abla}
\end{table}

\subsection{Case Study}
Figure ~\ref{fig:case} shows the comparison between the predictions of baseline methods and our MGAM model on three samples. As can be seen, in the first line, our model with multimodal fine-grained aligned information can detect the correct category ``Entertainment'' through the fine-grained element ``bookshelf'' of the image, which also demonstrates its effectiveness in boosting the model performance. Moreover, in the second line, we can see that there is no obvious mention of room facilities in the multimodal data. Our MGAM model has the ability to correct the wrong prediction made by BERT. These three examples further confirm our motivation that fine-grained integration is useful for multimodal data, so MGAM achieves better sentiment analysis performance.

\section{Conclusion}
In this paper, we propose the multimodal aspect category sentiment analysis task and the MACSA dataset, a Chinese multimodal aspect category sentiment analysis dataset with multimodal fine-grained aligned annotations. MACSA is the first dataset that contains fine-grained annotations for both image and text at the aspect category level. And we also propose the MACSA-hard dataset to study the aspect absent problem. On this basis, we verify the effect of the fine-grained aligned annotation through experiments on the multimodal ACSA task, propose the multimodal heterogeneous graph-based MGAM model, and report a series of benchmark models and results. The results demonstrate the superiority of our multimodal fine-grained fusion method on the multimodal aspect category sentiment analysis task.

\bibliographystyle{ACM-Reference-Format}
\bibliography{anthology}


\end{document}